\documentclass[10pt,twocolumn,letterpaper]{article}

\usepackage{wacv}
\usepackage{times}
\usepackage{epsfig}
\usepackage{graphicx}
\usepackage{caption}
\usepackage{subcaption}
\usepackage{amsmath}
\usepackage{amssymb}
\usepackage[normalem]{ulem}
\usepackage{booktabs}
\usepackage{array}
\usepackage{etoolbox}
\usepackage[table]{xcolor}
\usepackage{tabularx}
\usepackage[backref,breaklinks=true,bookmarks=false,colorlinks=true]{hyperref}
\usepackage{cleveref}
\usepackage{multirow}

\usepackage{pifont}
\newcommand{\cmark}{\ding{51}}%
\newcommand{\xmark}{\ding{55}}%
\newtheorem{prop}{Remark}

\newtheorem{subprop}{Remark}[prop]

\newcommand{\tableref}[1]{\autoref{table:sota_results}.\autoref{#1}}
\wacvfinalcopy 

\def\wacvPaperID{498} 

\ifwacvfinal\fi
\begin{document}

\title{Let there be a clock on the beach: \\ Reducing Object Hallucination in Image Captioning}

\author{Ali Furkan Biten  ~ ~ Lluis Gomez  ~ ~  Dimosthenis Karatzas\\
Computer Vision Center, UAB, Spain\\
{\tt\small {abiten, lgomez, dimos}@cvc.uab.es}
}

\maketitle
\ifwacvfinal\thispagestyle{empty}\fi

\begin{abstract}
Explaining an image with missing or non-existent objects is known as object bias (hallucination) in image captioning. This behaviour is quite common in the state-of-the-art captioning models which is not desirable by humans. To decrease the object hallucination in captioning, we propose three simple yet efficient training augmentation method for sentences which requires no new training data or increase in the model size. By extensive analysis, we show that the proposed methods can significantly diminish our models' object bias on hallucination metrics. 
Moreover, we experimentally demonstrate that our methods decrease the dependency on the visual features. All of our code, configuration files and model weights are available online\footnote{\url{https://github.com/furkanbiten/object-bias}}.
\end{abstract}
\section{Introduction}

In his seminal work~\cite{kuhn2012structure}, Kuhn stated that discoveries in anomalies usually lead to new paradigms. 
Machine Learning (ML) in its early days relied on hand-coded/crafted features (alas simple, elegant ones are favored) to create models. 
However, there was an anomaly that performed better than hand crafted features and it led to a paradigm shift 
called the Unreasonable Effectiveness of Data~\cite{halevy2009unreasonable},
where they simply advised ``to follow the data''.
Following the data was only the first step in the paradigm change, the second step was the amount of data. The introduction of big datasets like MSCOCO~\cite{lin2014microsoft} and ImageNet~\cite{deng2009imagenet} combined with the current advent in compute has resulted deep learning achieving significant feats~\cite{lecun2015deep}. Nevertheless, many works are published regarding the various failure cases and shortcuts that are exploited by deep models~\cite{geirhos2020shortcut}. 
These shortcuts can be found especially in Vision and Language tasks such as Image Captioning and Visual Question Answering (VQA) in the form of object hallucination~\cite{rohrbach2018object}, language prior~\cite{goyal2017making}, focusing on background~\cite{beery2018recognition}, spurious correlations~\cite{yang2020deconfounded}, action bias~\cite{yang2020deconfounded}, and gender bias~\cite{hendricks2018women}.
\begin{figure}[htp]
\begin{center}
\includegraphics[width=\linewidth,height=0.3\textwidth]{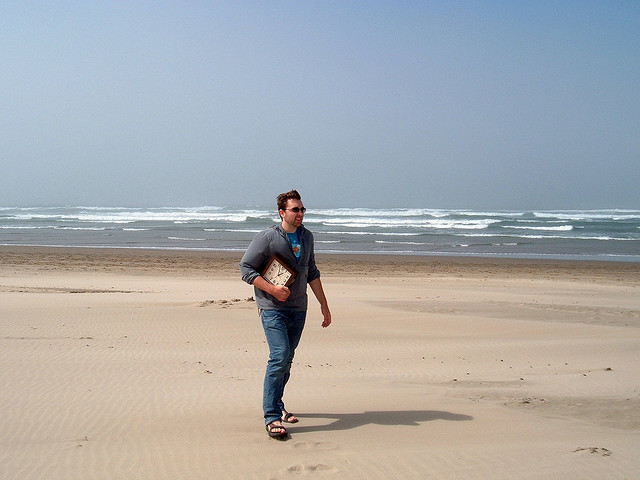}
\setlength\extrarowheight{3pt} 
\begin{tabularx}{\linewidth}{X}
  \fontfamily{lmss}\selectfont\textbf{UD}: A man on a beach with a surfboard \\
   \fontfamily{lmss}\selectfont\textbf{AoA}: A man standing on a beach holding a frisbee \\
   \fontfamily{lmss}\selectfont\textbf{Ours (UD)}: A man standing on a beach near the ocean \\
    \fontfamily{lmss}\selectfont\textbf{Ours (AoA)}: A man standing on a beach with a clock \\
\end{tabularx}

\caption{Standard approaches to image captioning are known to hallucinate on objects that do co-occur frequently, e.g. beach and frisbee or surfboard. Our method is capable of reducing object bias by normalizing the co-occurrence statistics, resulting in a reduction of hallucinated objects and the correct prediction of lower probability ones.}
\label{fig:teaser}
\end{center}
\vspace{-0.8cm}
\end{figure}

Solving the problem of object bias in image captioning is important for various reasons.
First and foremost, describing an image while failing to correctly identify objects is not desirable to humans~\cite{rohrbach2018object}. This is especially true for visually impaired people where they prefer correctness over coverage~\cite{macleod2017understanding} for obvious reasons. Secondly, even though the results of the captioning models are being pushed to the limit in automatic evaluation metrics, this does not translate to a decrease in object bias/hallucination~\cite{rohrbach2018object}. Finally, solving object hallucination is crucial for our models' generalization capabilities, allowing them to adapt easier to unseen domains. 

It is obvious that hallucination cannot be corrected by collecting even more data from the same biased world. The co-occurrence patterns will not change or they will be magnified. 
In other words, these biases do not seem to disappear neither with scaling up the dataset and nor with the increase in model size~\cite{geirhos2020shortcut}. 

In this work, we demonstrate that it is possible to reduce the object bias without needing more data or increase in the model size while not affecting the model's computational complexity and performance. 
More specifically, we tweak any existing captioning models by providing object labels as an additional input and employ a simple yet effective sampling strategy which consist of artificially changing the objects in the captions, e.g. modifying the sentence ``a \textit{person} is playing with a dog'' to ``a \textit{fork} is playing with a dog''. Along with a change in the sentence, in a corresponding way we also replace the object labels provided to the model. 

The reason is simple and can be traced back to co-occurrence statistics. By altering the co-occurrence statistics of the objects, we lessen the models' dependence on language prior and visual features as can be seen in Figure~\ref{fig:teaser}. 
Our contributions in this work are as follows:
\begin{itemize}
    \item  A simple method that can be applied to any captioning model to reduce object bias which requires no extra training data or increase in the model parameters.
    \item We improve the results on the hallucination metric CHAIR~\cite{rohrbach2018object} while obtaining a boost over our baseline models on image captioning evaluation metrics.
    \item We demonstrate that our technique works with two commonly used loss functions, cross entropy and REINFORCE~\cite{rennie2017self} algorithm.
\end{itemize}


\section{Related Work}
Following the advances of the encoder-decoder framework~\cite{Cho2014} with attention~\cite{Bahdanau2014} in machine translation, 
automatic image captioning took off using similar architectures~\cite{vinyals2015show, Xu2015}. The next advance in captioning came from using a pretrained object detector as feature extractor with two types of attention, top-down and bottom-up attention~\cite{anderson2017bottom}. In parallel, it was demonstrated that training captioning models with the REINFORCE algorithm~\cite{williams1992simple}, optimizing the evaluation metrics directly, had benefits over using cross-entropy loss~\cite{rennie2017self}. More recently, with the presentation of Transformers~\cite{vaswani2017attention}, a new family of models~\cite{huang2019attention} achieved state-of-the-art results. Image captioning recently shifted into new directions such as generating diverse descriptions~\cite{venugopalan2017captioning,deshpande2018diverse,xu2021towards} by allowing both grounding and controllability~\cite{cornia2019show, zhong2020comprehensive,chen2020say} while using various contextual information~\cite{biten2019good, tran2020transform}. 

Nevertheless, despite the continuous improvement on the classic captioning metrics, there are many biases exploited, that produce biases in the models. To compensate for gender bias where the models are known to prefer a certain gender over the other in specific settings,~\cite{hendricks2018women} proposed to tweak the original cross-entropy loss with confidence/appearance loss. 
Another bias in captioning is related to action bias where certain actions are preferred over others described by~\cite{yang2020deconfounded} where they employ causality~\cite{pearl2009causality} into the captioning models. More specifically, their proposed method uses 4 layers LSTM with running expected average on ConceptNet~\cite{liu2004conceptnet} concepts for each word produced in the caption, which it introduces a significant computation overload. Similarly, \cite{agarwal2020towards} modifies the images with generative models to reduce the effect of spurious correlations in Visual Question Answering task.

\cite{rohrbach2018object} show that contemporary captioning models are prone to object bias. Moreover, they describe that evaluation metrics merely measure the similarity between the ground truth and produced caption, not capturing image relevance. Consequently, they propose two metrics to  quantify the degree of hallucination of objects, namely CHAIRs and CHAIRi. CHAIR metrics evaluates how much our models produced wrong object labels at sentence level (hence CHAIRs) and at object level (hence CHAIRi). 
Surprisingly, the object hallucination problem has not received the attention it deserves. 
In this work, we try to diminish object bias without enlarging the model size or using extra data. We do so following a simple strategy that can be used with any model that accepts object detection features as inputs.
\begin{figure*}
    \centering
    \includegraphics[width=\textwidth]{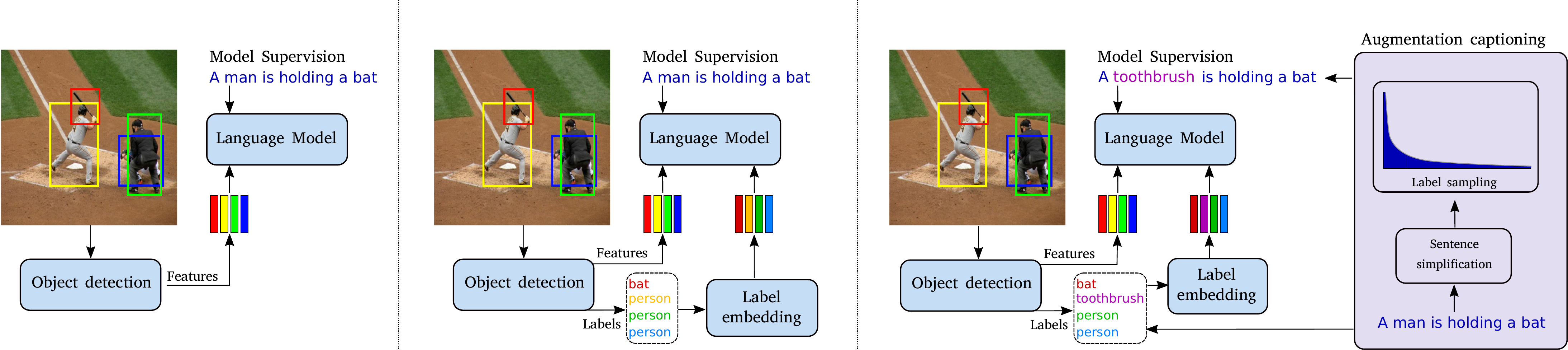}
    \caption{Most current models for image captioning utilize object-level visual features extracted from an object detection network (left diagram). In this paper we propose a simple tweak that consists of providing also the object labels as input (center diagram). The concatenation of label embeddings to visual features allows us to employ data augmentation techniques on the object labels and model supervision (captions) to fix the object bias in our models (right diagram).}
    \label{fig:main}
\end{figure*}

\section{Methods}
As mentioned previously, we try to reduce the object bias that exists inherently in existing models. The main cause of object bias is the systematic co-occurence of specific object categories in images of our training datasets, we therefore hypothesize that making the co-occurence statistics matrix more uniform will make our models hallucinate less. Accordingly, we devise a series of data augmentation techniques to achieve this goal.

\subsection{A Small Tweak to Any Captioning Model}
We would like to start off by giving a concise and general introduction to models in image captioning. After the introduction of top-down bottom-up attention~\cite{anderson2017bottom}, most of existing models for image captioning utilize object-level visual features extracted from an object detection network. More formally, given an image $I$, a set of bounding box features $V = \{v_1, v_2, ..., v_n\}$ are obtained by passing it through a pretrained object detector $\mathcal{O}$, \ie $V = \mathcal{O}(I)$. These features are combined with an attention mechanism to be later fed into Language Models ($\mathcal{L}$) to generate a sentence $S = \{w_1, w_2, ..., w_k\}$ where most common variants of $\mathcal{L}$ are Transformers~\cite{vaswani2017attention} and LSTM~\cite{hochreiter1997long}:
\begin{equation}
\begin{split}
    \bar{V} &= Att(V, h) \\
    P(w_j), \ h &= \mathcal{L}(w_j|\bar{V}, w_1, w_2, ..., w_{j-1})
\end{split}
\label{eq:captioning}
\end{equation}
This formulation can be seen more clearly on the left part of Figure~\ref{fig:main}.

Our tweak to the aforementioned formulation is to simply concatenate the object labels found in an image with bounding box features (middle part of the Figure~\ref{fig:main}). More formally, we extend the set of bounding box features from $V$ to $\Tilde{V} = \{v_1, v_2, ..., v_n, l_1, l_2, ..., l_i\}$ where $l_i$ is the $i^{th}$ embedded object label. After the concatenation, we replace $V$ with $\bar{V}$ and follow exactly the same training procedure outlined in~\Cref{eq:captioning}. 

Concatenation of label embeddings to visual features allows us to employ our data augmentation techniques. Since we use the labels as input to our models, we can directly alter them as we see fit. In the following sections, we describe the strategy behind the augmentation of labels.

\subsection{Sentence Simplification}
A first step to all our data augmentation methods is sentence simplification. By sentence simplification we refer to removing adjectives that are used in the captions for objects in the scene. As an example, we would like to modify the sentence ``A small black cat is sitting on top of an old table'' into ``A cat is sitting on top of a table.''. The reasons are twofold, one of which is that there are adjectives that can not hold true for every object, e.g. ``small'' and ``black'' can be used for a cat but this will not be correct when cat is artifically changed with another object such as elephant or banana. Secondly, simplifying the sentence in this way provides another variation of sentences, acting as type of a regularizer for captioning models to exploit language prior existing in the dataset. 

To achieve this goal, we first analyze every caption with a Part-Of-Speech (POS) and find all the noun phrases corresponding to sentences. However, these noun phrases do not necessarily have to refer to objects found in an image. That is why, we make use of synonyms list for object classes that exist in the dataset (e.g. 80 objects in MSCOCO) and filter the noun phrases that include the object name or its synonyms. As a final step, we replace the whole noun phrase with the root of the phrase.

\subsection{Augmentation of Sentences}
After simplifying the sentences, we employ different sampling strategies to pick which object to replace. More formally, given a sentence containing objects $o_i$ and $o_j$, we sample object $o_k$ to replace $o_j$ according to the distribution $P(o_k|o_i)$. Now, we explain in detail which distributions we use to augment the sentences.

\subsubsection{Uniform Sampling}
The choice of uniform sampling is inspired by our hypothesis on creating  a uniform object label co-occurrence matrix.
In its most simplest form, we make use of uniform distribution for sampling, where 
\begin{equation}
    P(o_k|o_i) = P(o_k) = 1/N.
\end{equation}
In other words, every object has an equal probability to be sampled where dataset statistics are disregarded. 
The next two distribution takes into account the discarded dataset statistics.  
\subsubsection{Inverse Multinomial Sampling}
The most accessible statistics one can obtain regarding any given dataset is the co-occurrence matrix $M \in R^{N \times N}$ where $M_{ij}$ refers to the co-occurrence statistics of objects $o_i$ and $o_j$ and $N$ is the number of objects. 
We define a new distribution which considers dataset statistics called inverse multinomial by making use of $M$ where
\begin{equation}
    P(o_k|o_i) = \frac{1}{\Tilde{M}_{ik}} \ \text{where}\ \Tilde{M}_{ik} = \frac{M_{ik}}{\sum_k{M_{ik}}}
\end{equation}

With inverse multinomial, we sample object $o_k$ if the occurence is low with object $o_i$. On the other hand, if object $o_k$ and $o_i$ co-occurs frequently in the dataset, then the probability of selecting $o_k$ will be quite low.  

\subsubsection{Updating Co-Occurence Matrix}
Although inverse multinomial sampling increases the chance of low frequency pairs to be sampled, it prevents creating a new bias for low frequency pairs. To circumvent the problem, we determine to keep track of the matrix $M$ and constantly update according to the sampled pair. More formally, the distribution is defined as:
\begin{equation}
\begin{split}
    P(o_k|o_i) &= \frac{1}{\Tilde{M}_{ik}} \ \text{where}\ \Tilde{M}_{ij} = \frac{M_{ij}}{\sum_j{M_{ij}}} \\
    M_{ik} &= M_{ik} + 1, M_{ij} = M_{ij} - 1 
\end{split}
\end{equation}
By keeping track of co-occurence statistics in training diminishes the prospect of models finding a shortcut as well as allowing faster convergence to a uniform $M$.

\section{Experiments}
\subsection{Dataset and Baseline Models}

\textbf{MSCOCO:}~\cite{lin2014microsoft}. We use the most commonly used captioning dataset,  MSCOCO~\cite{lin2014microsoft}. We follow the literature on using the `Karpathy' split~\cite{Karpathy}. The split contains 113,287 training images with 5 captions each and 5k images for validation and testing. 

\textbf{Evualation Metrics:} To evaluate caption quality, we report the standard automatic evaluation metrics; CIDEr~\cite{vedantam2015cider}, BLEU~\cite{Papineni2002}, METEOR~\cite{denkowski2014meteor}, 
SPICE~\cite{anderson2016spice}. Moreover, we include the newly introduced metric called SPICE-U~\cite{wang2020towards} which is a variant of SPICE where it rewards for uniqueness of sentences. Finally, we provide the hallucination metrics CHAIRs~\cite{rohrbach2018object} and CHAIRi~\cite{rohrbach2018object} for sentence and object level, respectively. In CHAIR metrics, lower is better.

\textbf{UpDown (UD):}~\cite{anderson2017bottom}. The bottom-up and top-down attention model utilizes the salient image regions proposed by object detector pretrained on VG~\cite{krishnavisualgenome} and then weighting the regions by employing an attention mechanism calculated according to Language Models' hidden state. 

\textbf{AoA:}~\cite{huang2019attention}.  The attention on attention model extends the conventional Transformers~\cite{vaswani2017attention} model by including another attention to determine the relevance between attention results and queries.
When we train with object labels given as inputs, we refer those models as UD-L and AoA-L. 

\subsection{Implementation Details}
All our models are implemented on top of publicly available code\footnote{\url{https://github.com/ruotianluo/self-critical.pytorch}}. We use Adam~\cite{kingma2014adam} optimizer with batch size 10 and learning rate 0.0002 and 0.0005 for UpDown~\cite{anderson2017bottom} and AoA~\cite{huang2019attention}, respectively. Both models are trained for 30 epochs and we kept the best models according to best score on validation set on Cider-D~\cite{vedantam2015cider}. We generate sentences with no beam search and both models use visual features provided by~\cite{anderson2017bottom}. For embedding the object labels, we utilize FastText~\cite{joulin2016fasttext}.

\newcounter{rowno}
\setcounter{rowno}{-2}
\definecolor{gray}{RGB}{210,210,210}
\newcolumntype{a}{>{\columncolor{gray}}c}

\begin{table*}[htp]
\caption{Results of image captinoning models on Karpathy test split. * numbers are provided by ~\cite{rohrbach2018object} with beam search 5. B-4: Bleu-4, M: Meteor, C: Cider, S: Spice, S: Spice-U, CHs: CHAIRs, CHi: CHAIRi, UD: UpDown, AoA: Attention on Attention, Uni: Uniform Sampling, Inv: Inverse Multinomial Sampling,  Occ: Co-occurence Updating. In CHAIR metrics, lower is better.}
\centering
\resizebox{\textwidth}{!}{\begin{tabular}
{
@{\makebox[2em][r]{\refstepcounter{rowno}\label{row:\therowno}\ifnum\value{rowno}<1\else\thetable.\therowno\space\fi}}
|c|cccc aac|ccccc aac}
\toprule

&                \multicolumn{7}{c}{\textbf{Cross Entropy}}             &  & \multicolumn{7}{c}{\textbf{Self Critical}} 

\\ 
\hline
\midrule
\textbf{Model}           & \multicolumn{1}{c}{\textbf{B-4} $\uparrow$} & \multicolumn{1}{c}{\textbf{M} $\uparrow$} & \multicolumn{1}{c}{\textbf{C} $\uparrow$} & \multicolumn{1}{c}{\textbf{S} $\uparrow$} & \multicolumn{1}{c}{\textbf{CHs} $\downarrow$} & \multicolumn{1}{c}{\textbf{CHi} $\downarrow$} & \multicolumn{1}{c}{\textbf{S-U} $\uparrow$} &  & \multicolumn{1}{c}{\textbf{B-4} $\uparrow$} & \multicolumn{1}{c}{\textbf{M} $\uparrow$} & \multicolumn{1}{c}{\textbf{C} $\uparrow$} & \multicolumn{1}{c}{\textbf{S} $\uparrow$} & \multicolumn{1}{c}{\textbf{CHs} $\downarrow$} & \multicolumn{1}{c}{\textbf{CHi} $\downarrow$} & \multicolumn{1}{c}{\textbf{S-U} $\uparrow$} \\
\midrule

UD-VC~\cite{wang2020visual}  & 39.5&	29&		130.5& -& 10.3  & 6.5	& -	& &- &-&	-&	-&	-&	-& - \\

AoA-VC~\cite{wang2020visual}  & 39.5&	29.3&		131.6& -&8.8  & 5.5	& -	& &- &-&	-&	-&	-&	-& - \\

UD-DIC~\cite{yang2020deconfounded}  & 38.7&	28.4&	128.2&	21.9&	10.2&	6.7& -	& &- &-&	-&	-&	-&	-& - \\

UD-MMI~\cite{wang2020towards}  & 22.77&	28.84&	106.42&	20.72&	7.8&-&	25.27&  &- &-&	-&	-&	-&	-& - \\

AoA-MMI~\cite{wang2020towards}  & 27.18&	30.39&	128.15&	22.81&	9.28&-&	26.53	& &- &-&	-&	-&	-&	-& - \\

DiscCap~\cite{wang2020towards}  & 21.58&	27.42&	110.9&	20.27&	10.84&  -	&	24.52& &- &-&	-&	-&	-&	-& - \\

LRCN~\cite{donahue2015long}* & - & 23.9 &90.8 &17.0 & 17.7 & 12.6& - & & - &23.5 & 93.0 & 16.9 & 17.7 & 12.9 & - \\

FC~\cite{rennie2017self}*  & - & 24.9 & 95.8 &	17.9 &	15.4&	11& -	& &- &25&	103.9&	18.4&	14.4&	10.1& - \\

Att2In~\cite{rennie2017self}* & - & 25.8&	102&	18.9&	10.8&	7.9& -	& &- &25.7&	106.7&	19&	12.2&	8.4& - \\

UD~\cite{anderson2017bottom}* & - & 27.1&	113.7&	20.4&	8.3&	5.9	& -	& &- &27.7&	120.6&	21.4&	10.4&	6.9& - \\

NBT~\cite{lu2018neural}* & - & 26.2& 105.1&19.4& 7.4& 5.4	& -	& &- &-&	-&	-&	-&	-& - \\

GAN~\cite{shetty2017speaking}* & - & 25.7&	100.4&	18.7&	10.7&	7.7	& -	& & - &-&	-&	-&	-&	-& - \\

\hline

 {UD}         & 33.2                              & 26.9                              & 108.4                            & 20.0                             & 10.1                                & 6.9                                 & 24.05                              &  & 36.5                              & 27.8                              & 121.5                            & 21.3                             & 11.9                                & 7.7                                 & 23.85                              \\
{UD-L}          & 34.4                              & 27.3                              & 112.7                            & 20.7                             & 6.4                                 & 4.1                                 & {24.68}                     &  & 37.7                              & 28.6                              & 124.7                            & 22.1                             & 5.9                                 & {3.7}                        & 25.41                              \\
{UD-L + Uni}                          & 34.2                              & 27.2                              & 112.4                            & 20.6                             & 6.3                                 & 4.0                                   & 24.61                              &  & 37.6                              & 28.7                              & 125.2                            & {22.3}                    & \textbf{{5.8}}                        & \textbf{{3.7}}                        & 25.54                              \\
{UD-L + Inv}                           & {34.3}                     & {27.3}                     & {112.6}                   & {20.7}                    & 6.2                                 & 4.0                                   & 24.05                              &  & {37.8}                     & 28.7                              & {125.4}                   & {22.3}                    & 5.9                                 & 3.8                                 & {25.60}                     \\
{UD-L + Occ}                          & 33.9                              & 27.0                              & 110.7                            & 20.3                             & \textbf{{5.9}}                        & \textbf{{3.8}}                        & 24.52                              &  & 37.7                              & 28.7                              & 125.2                            & 22.2                             & \textbf{{5.8}}                        & \textbf{{3.7}}                        & 25.58                              \\
\hline

{AoA} & 33.7                              & 27.4                              & 111.0                            & 20.6                             & {9.1}                        & 6.2                                 & 24.57                              &  & {38.8}                     & {28.7}                     & {127.2}                   & {22.4}                    & 9.6                                 & 6.1                                 & 24.68                              \\
{AoA-L} & 33.1                              & 27.0                              & 110.0                            & 20.3                             & 7.1                                 & 4.4                                 & 24.30                              &  & 35.9                              & 28.0                              & 119.6                            & 21.7                             & 7.8                                 & 4.8                                 & 24.81                              \\
{AoA-L + Uni}              & 34.1                              & 27.2                              & 111.4                            & 20.5                             & \textbf{{6.2}}                        & \textbf{{3.9} }                       & 24.58                              &  & 35.1                              & 27.8                              & 117.7                            & 21.4                             & 7.3                                 & 4.5                                 & 24.58                              \\
{AoA-L + Inv}                 & {34.3}                     & {27.3}                     & {112.0}                   & {20.6}                    & 6.5                                 & 4.1                                 & {24.93}                     &  & {35.7}                     & {28.0}                     & {119.2}                   & {21.8}                    & 7.5                                 & 4.6                                 & {24.93}                     \\
{AoA-L + Occ}                 & {34.3}                     & 27.1                              & 111.3                            & 20.5                             & \textbf{{6.2} }                       & \textbf{{3.9}}                        & 24.57                              &  & 34.5                              & 27.5                              & 116.0                            & 21.1                             & \textbf{7.0}                          & \textbf{4.3}                        & 24.20                              \\ \bottomrule
\end{tabular}}
\label{table:sota_results}
\end{table*}
We use both of the commonly used training losses employed by the literature, namely cross-entropy and REINFORCE~\cite{rennie2017self}. For every variant of our model, we randomly choose to use original sentences or augmented sentences according to flip of a coin as ground-truth. All the models trained with our augmentation are fine-tuned to allow faster convergence and to see if we can reduce the ``learned'' biases of our models. Finally, we always use the ground truth object labels as input to our models and use X101-FPN from Detectron2~\cite{wu2019detectron2} library to obtain object labels for testing.  All the code, model weights and configuration file necessary for the hyper-parameters will be released upon accceptance. 

\subsection{Comparison to State of Art}

We present the results of our models as well as the state-of-the-art model results in~\autoref{table:sota_results}. 
First and foremost, UD-VC ad AoA-VC (row \tableref{row:1}, \tableref{row:2}) uses the features extracted from state of the art object detector while concatenating with the original features provided by UpDown~\cite{anderson2017bottom}, \ie they use 2 FasterRCNN architecture in their model training. While UD-DIC (row \tableref{row:3}) uses 4 deep LSTMs~\cite{hochreiter1997long} to find matching between produced words and ConceptNet~\cite{liu2004conceptnet} labels. Moreover, UD-MMI (\tableref{row:4}) and AoA-MMI (\tableref{row:5}) trains an LSTM without any visual features to detect the common and not unique sentences and later to be used at inference time. From aforementioned models, we observe that beating state-of-the-art results or increase in the model size or even using better features does not result in our models hallucinating less.

\begin{prop}
Increase in the model size (parameters) or boost in the image captioning metrics does not result in decrease in CHAIR metrics. 
\end{prop}

Subvariant of this conclusion can be also seen in REINFORCE~\cite{rennie2017self} training. It is common practice in captioning community to train the models first with cross entropy and then with self-critical loss~\cite{rennie2017self} on CIDER-D~\cite{vedantam2015cider}. While this training ensures a significant boost on the automatic metrics especially on CIDER, it makes our models hallucinate more (can be seen in row \tableref{row:7}, \tableref{row:8}, \tableref{row:11}, \tableref{row:13} and \tableref{row:18}). 

\begin{subprop}
Self-Critical training leads to increase in the captioning metrics while making models hallucinate more.
\end{subprop}

The next point we would like to move to is regarding our methods starting from~\tableref{row:13}. Simply by adding the object labels as input we notice an improvement on CHAIR metrics for both of the models. This progress can also be observed on classic image captioning metrics for UpDown model. 
Furthermore, we note that addition of labels also reaches the reported numbers in~\tableref{row:10} while significantly diminishing the object bias on both sentence and object level. Finally, we see that this simple technique of concatenating object labels and visual features already gives state-of-the-art results in object hallucination by around 1 to 4\%. 

\begin{table*}[t]
\caption{Results on Karpathy Test split. The numbers are obtained by using ground truth object labels instead of using object detector.}
\label{table:gt}
\centering
\resizebox{\textwidth}{!}{\begin{tabular}{cc|cccccc|cccccc}

\toprule
&                                           & \multicolumn{6}{c}{\textbf{Cross Entropy}} &  \multicolumn{6}{c}{\textbf{Self Critical}}                                                              \\

\midrule
\textbf{Model} &  \textbf{Aug}                   & \multicolumn{1}{c}{\textbf{Bleu-4} $\uparrow$} & \multicolumn{1}{c}{\textbf{METEOR} $\uparrow$} & \multicolumn{1}{c}{\textbf{CIDEr} $\uparrow$} & \multicolumn{1}{c}{\textbf{SPICE} $\uparrow$} & \multicolumn{1}{c}{\textbf{CHAIRs} $\downarrow$} & \multicolumn{1}{c}{\textbf{CHAIRi} $\downarrow$} &
\multicolumn{1}{c}{\textbf{Bleu-4} $\uparrow$} & \multicolumn{1}{c}{\textbf{METEOR} $\uparrow$} & \multicolumn{1}{c}{\textbf{CIDEr} $\uparrow$} & \multicolumn{1}{c}{\textbf{SPICE} $\uparrow$} & \multicolumn{1}{c}{\textbf{CHAIRs} $\downarrow$} & \multicolumn{1}{c}{\textbf{CHAIRi} $\downarrow$} \\ 
\midrule
{\textbf{UD}}    &   - & \textbf{34.6}                     & \textbf{27.4}                     & {112.9}                   & {20.8}                    & {4.5}                        & {2.8}                        & 37.9                              & {28.7}                     & {125.9}                   & {22.3}                    & \textbf{3.5}                        & \textbf{2.2}                        \\
{\textbf{UD}}    &            U                              & \textbf{34.6}                     & \textbf{27.4}                     & 113.4                            & 20.8                             & 4                                   & 2.5                                 & 38.0                              & 28.9                              & 126.2                            & 22.5                             & 3.7                                 & 2.3                                 \\
\textbf{UD}    &           IM                             & 34.5                              & \textbf{27.4}                     & \textbf{114.0}                   & \textbf{20.9}                    & 3.9                                 & 2.4                                 & \textbf{38.0}                              & \textbf{28.8}                              & \textbf{126.4}                            & \textbf{22.5}                             & 3.9                                 & 2.4                                 \\
\textbf{UD}    &            Occ                            & 34.0                              & 27.1                              & 111.6                            & 20.5                             & \textbf{3.6}                        & \textbf{2.2}                        & \textbf{38.0}                     & \textbf{28.8}                     & \textbf{126.4}                   & \textbf{22.5}                    & \textbf{3.5}                        & \textbf{2.1}                        \\
\midrule
\textbf{AoA}   &            -          & 33.4                              & 27.2                              & 111.4                            & 20.5                             & 4.4                                 & 2.7                                 & 36.2                              & 28.3                              & 121.3                            & 22.0                             & 4.3                                 & 2.6                                 \\
\textbf{AoA}   &            U                              & 34.4                              & 27.3                              & 112.5                            & 20.7                             & \textbf{2.7}                        & \textbf{1.6}                        & 35.5                              & \textbf{28.0}                     & \textbf{119.2}                   & \textbf{21.7}                    & {3.9}                        & 2.3                                 \\
\textbf{AoA}   &           IM                             & \textbf{34.6}                     & \textbf{27.4}                     & \textbf{113.4}                   & \textbf{20.8}                    & 3.1                                 & 1.9                                 & \textbf{36.1}                     & \textbf{28.3}                     & \textbf{121.0}                   & \textbf{22.0}                    & 3.9                                 & 2.3                                 \\
\textbf{AoA}   &            Occ                            & 34.4                              & \textbf{27.4}                     & 113.0                            & 20.7                             & \textbf{2.7}                        & \textbf{1.6}                        & 34.9                              & 27.7                              & 117.4                            & 21.3                             & \textbf{3.7}                        & \textbf{2.2}\\
\bottomrule
\end{tabular}}
\end{table*}
\begin{prop}
Merely concatenating the labels with visual features results in the decline of hallucination of our models while beating state-of-the-art models on CHAIR metrics.
\end{prop}

Before we focus on our augmentation techniques, we want to point out that reducing object hallucination from 10\% to 6\% is not an equivalent to reducing it from 6\% to 2\%. The reason is that there are 2 different elements affecting the hallucination, one of which is the dataset bias which what we are trying to solve and the other one is the noisy and incorrect FasterRCNN features. From the next section, we see that our methods upper bound is around 2-3\%, suggesting that the rest of the hallucination is mostly coming from the visual features. That said, it can be seen that we even better the results of row~\tableref{row:14} and row~\tableref{row:19} in comparison to our proposed techniques by around 0.5 to 1\%. 

\begin{prop}
We demonstrate that our proposed techniques can reduce the object bias on the same model architectures. 
\end{prop}

Furthermore, we remark that we always obtain the best results on CHAIR metrics with the Co-Occurrence Updating technique although we usually obtain a decrease in the other common metrics. We also see that Inverse Multinomial sampling results in the best performance in the classic captioning metrics. Moreoever, Co-Occurence Updating always achieves the best CHAIR scores out of all the different sampling.

\begin{subprop}
Inverse Multinomial achieves the best scores on standard captioning metrics while Co-Occurence updating performs the best on CHAIR metrics. 
\end{subprop}

Finally, we report the recently introduced metric SPICE-U~\cite{wang2020towards} where it evaluates how unique and informative a caption is. We take interest in the said metric since our concern was that proposed augmentation can make captioning models produce more repetitive or less informative captions because of the sentence simplification. As can be observed from the~\autoref{table:sota_results}, even in the cases where we have a drop on standard image captioning metrics, we still improve on SPICE-U. In row~\tableref{row:14}-\tableref{row:17}, we even have 2\% improvement in self-critical training. This is quite encouraging especially compared to SOTA numbers on row~\tableref{row:4}-\tableref{row:6} where we even beat the numbers without the need of training an extra LSTM.

\begin{prop}
Our techniques can improve or at least stay the same as the base model on producing informative and unique captions.
\end{prop}


\subsection{What if we have perfect label extractor?}

As the title reads, we try to figure out the upper bound for our techniques. In other words, since it is known that object detectors are far from providing the perfect labels, we test our methods with the ground truth annotations of object labels to see the full performance of our different methods, given in Table~\ref{table:gt}. We use the same models provided in Table~\ref{table:sota_results}.

First conclusion is that we see an improvement on all the metrics with the usage of ground truth. This is quite expected since we have trained with the ground truth annotations. 

\begin{prop}
With the perfect object detector, we can improve on all the metrics.
\end{prop}

One important remark is that the gap between the models with labels and models trained with our augmentation is much bigger. Particularly, for UpDown we see the gap becomes 0.9\% and 0.6\% while for AoA, it is 1.7\%, 1.1\% on CHAIRs and CHAIRi, respectively. This suggests that our proposed augmentation will reach even higher values with the advances on object detector's performance. 

\begin{subprop}
Our proposed methods can achieve higher performance simply by obtaining more precise labels.
\end{subprop}

Finally, it can be appreciated that in all of the models whether trained with cross-entropy or self-critical, Co-Occurrence Updating always accomplishes the best scores on CHAIR metrics, confirming our hypothesis on creating a uniform co-occurence matrix causing a decrease on object bias.

\begin{subprop}
By making the co-occurrence matrix uniform causes our models to have \textbf{the least} object bias.
\end{subprop}

\begin{figure*}

\begin{subfigure}{.5\textwidth}
  \centering
  \caption{CHAIRs scores}
  \vspace{-3em}
  \label{fig:chairs}
  \includegraphics[width=1.\linewidth]{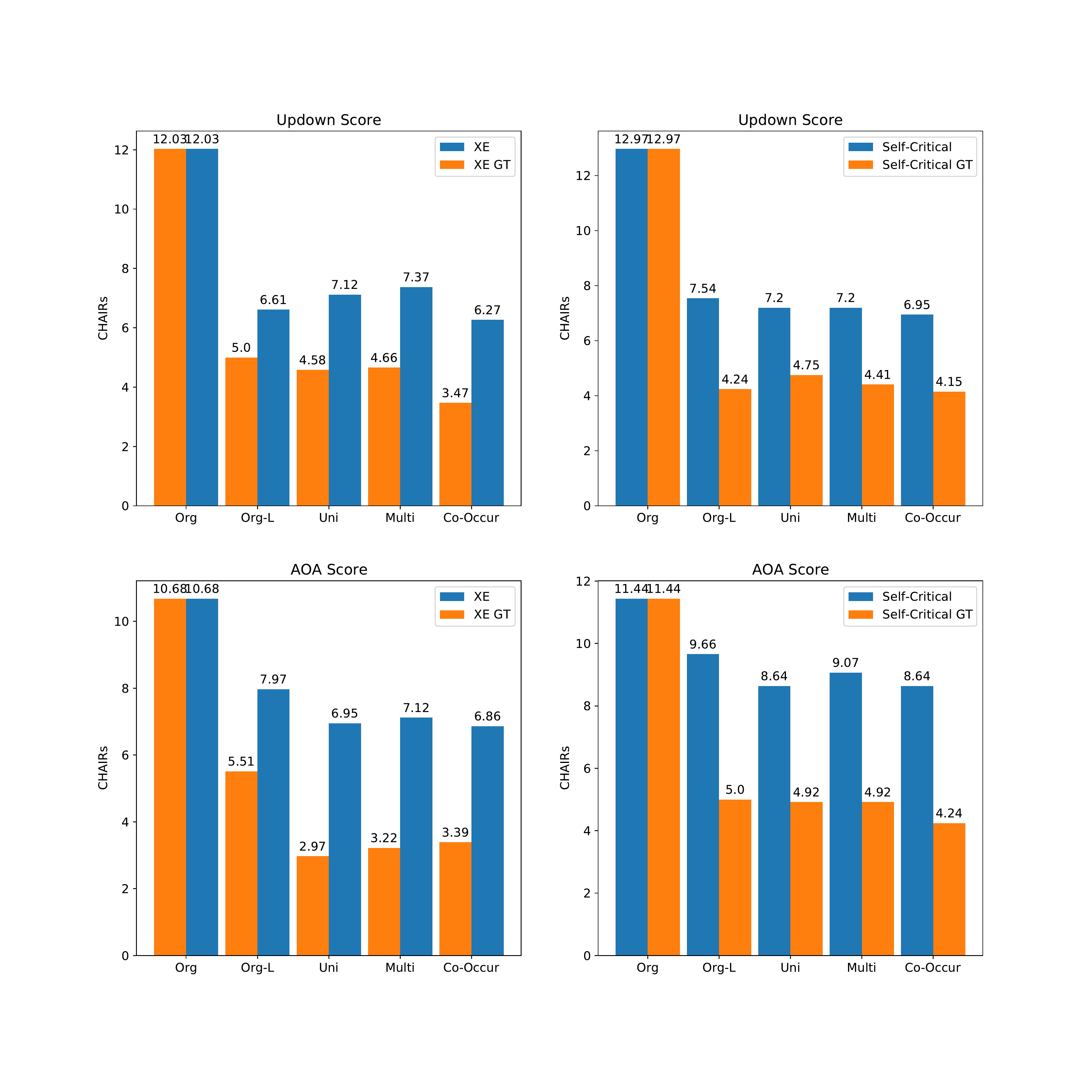}
\end{subfigure}%
\begin{subfigure}{.5\textwidth}
  \centering
  \caption{CHAIRi scores}
  \vspace{-3em}
  \label{fig:chairi}
  \includegraphics[width=1.\linewidth]{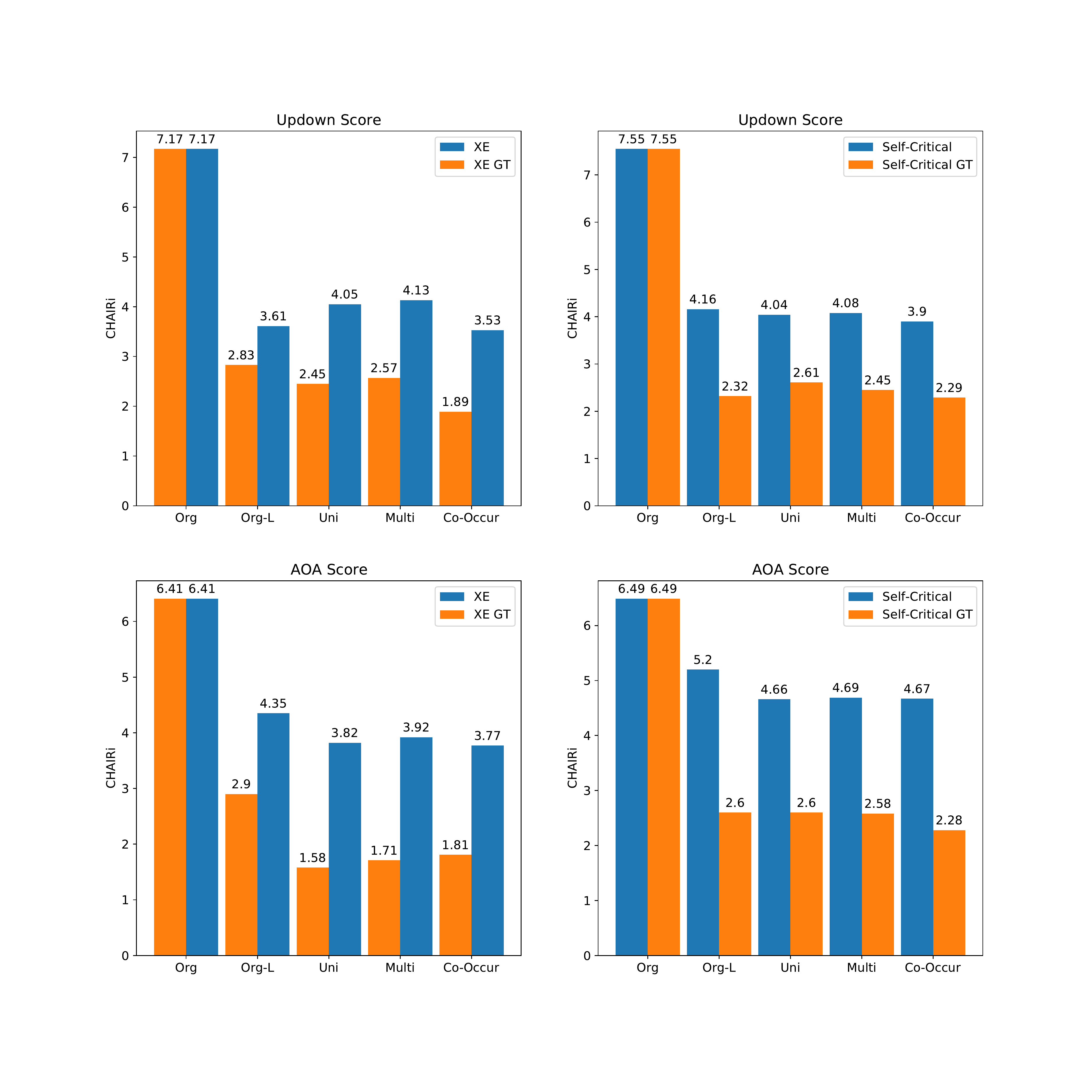}
\end{subfigure}
\vspace{-3em}
\caption{Bar plot on low frequency pairs. We provide all the models we trained with object detector labels and ground truth labels. We select the sentences which contain objects pairs that has less than 200 co-occurrence. }
\label{fig:low_freq}
\end{figure*}

\begin{table}[htp]
\centering
\resizebox{\columnwidth}{!}
{\begin{tabular}{c|cc|cc|cc}
 &  &  & \multicolumn{2}{c}{\textbf{FRCNN}} & \multicolumn{2}{c}{\textbf{Ground Truth}} \\
{} & \textbf{Vis Feat} & \textbf{Labels} & \multicolumn{1}{c}{\textbf{CHAIRs}} & \multicolumn{1}{c}{\textbf{CHAIRi}} & \multicolumn{1}{c}{\textbf{CHAIRs}} & \multicolumn{1}{c}{\textbf{CHAIRi}} \\
\hline
\textbf{UD-L} & {\cmark} & {\xmark} & {9.2} & {6.6} & {-} & {-} \\
\textbf{UD-L + Uni} & \cmark & \xmark & {9.4} & {6.7} & - & - \\
\textbf{UD-L + Inv} & \cmark & \xmark & 9.2 & {6.6} & {-} & {-} \\
\textbf{UD-L + Occ} & \cmark & \xmark & 9.8 & 7.1 & - & - \\
\hline
\textbf{UD-L} & \xmark & \cmark & 35.8 & 29.1 & {35.7} & {28.7} \\
\textbf{UD-L + Uni} & \xmark & \cmark & 26.1 & 18.8 & {24.7} & {17.3} \\
\textbf{UD-L + Inv} & \xmark & \cmark & {29.2} & {21} & {{28}} & {{19.8}} \\
\textbf{UD-L + Occ} & \xmark & \cmark & 20.2 & {13.6} & {17.1} & {11.2}
\end{tabular}}
\caption{Results on Karpathy Test split. We either provide to our models only visual features or object label embeddings.}
\label{table:zero_out}
\end{table}
\subsection{Data augmentation effect on the models} 
Our next set of experiments is done to find out what is the proposed augmentation provides to the model. To take a stab at the problem, we decided to zero out either the visual features or the object labels at inference time to see how much importance they have on hallucination. Our numbers can be seen in Table~\ref{table:zero_out}. Primarily, we appreciate that the results are much better when using visual features than when using object labels. This is anticipated and can be thought as taking away the ``eyes'' of the model. However, we identify that visual features holds more significance for UD-L than the models trained with the augmentation (UD-L+Occ and UD-L+Uni). 

\begin{prop}
The proposed training leads to models to put more emphasis on the labels while reducing the dependence on visual features. 
\end{prop}

Moreover, it can be appreciated that our model trained with Co-Occurence Updating puts less importance on the visual features or utilizes it to a lesser degree than the other models. 
This point is especially reinforced when we examine the zeroing out the visual features. We recognize that models trained with our augmentations utilizes much more the provided labels, in which from UD-L to UD-L+Occ, there is a 15\% improvement. Another evidence for the statement is that in UD-L from object detection labels to ground truth, there is simply 0.1\%, 0.4\% improvement. Furthermore, we can even see that this gap grows even more when the ground truth is used as input to our models. We notice that the same difference when used ground truth raises to 18\% and 17\% for CHAIRs and CHAIRi, respectively. 

\begin{subprop}
Co-Occurrence Updating exploits the labels the most out of other 3 models. 
\end{subprop}
\begin{figure*}[h]
    \centering
    \begingroup
    \setlength{\tabcolsep}{2pt} 
    \renewcommand{\arraystretch}{1} 
    \setlength{\fboxsep}{1pt}
    \begin{tabular}{c c c c c}
    {\includegraphics[width=0.19\textwidth, height=0.19\textwidth]{img/COCO_val2014_000000460682.jpg}} &
    {\includegraphics[width=0.19\textwidth, height=0.19\textwidth]{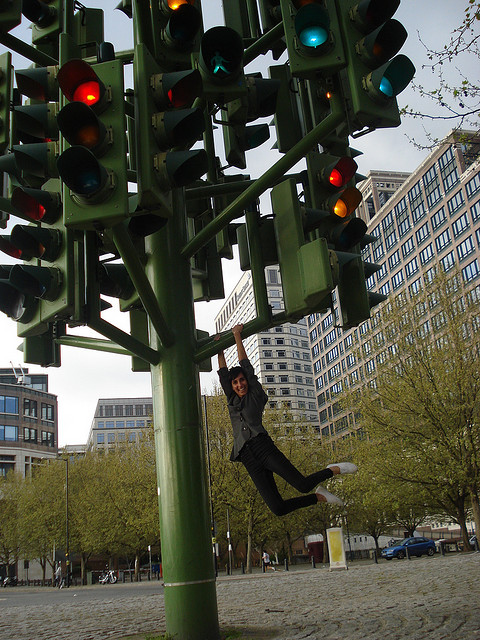}} &
    {\includegraphics[width=0.19\textwidth, height=0.19\textwidth]{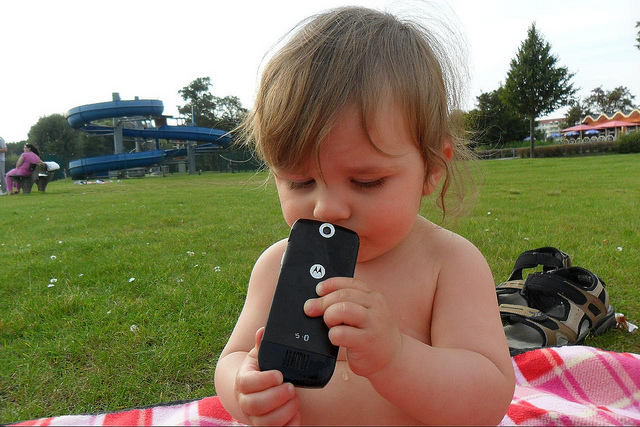}} &
    {\includegraphics[width=0.19\textwidth, height=0.19\textwidth]{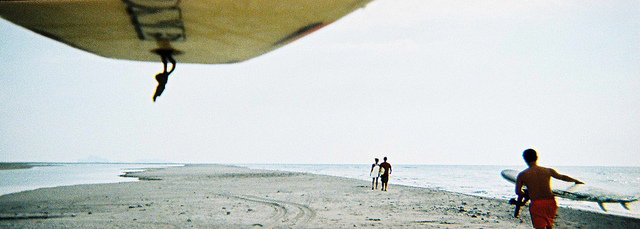}} &
    {\includegraphics[width=0.19\textwidth, height=0.19\textwidth]{img/COCO_val2014_000000006393.jpg}} \\
  \multicolumn{1}{|p{0.19\textwidth}|}{\footnotesize {\fontfamily{lmss}\selectfont \textbf{UD}: A dog is sitting in the grass near a lake}} &
  \multicolumn{1}{|p{0.19\textwidth}|}{\footnotesize {\fontfamily{lmss}\selectfont \textbf{UD}: A man is jumping a street on a skateboard}}     &
  \multicolumn{1}{|p{0.19\textwidth}|}{\footnotesize {\fontfamily{lmss}\selectfont \textbf{UD}: A young child holding a remote control in his hand}}     &
  \multicolumn{1}{|p{0.19\textwidth}|}{\footnotesize {\fontfamily{lmss}\selectfont \textbf{UD}: A group of people on a beach with a kite}}  &
  \multicolumn{1}{|p{0.19\textwidth}|}{\footnotesize {\fontfamily{lmss}\selectfont \textbf{UD}: A woman is looking at a cell phone}} \\

    \multicolumn{1}{|p{0.19\textwidth}|}{\footnotesize {\fontfamily{lmss}\selectfont \textbf{AoA}: A dog running in a field near a body of water}} &
  \multicolumn{1}{|p{0.19\textwidth}|}{\footnotesize {\fontfamily{lmss}\selectfont \textbf{AoA}: A man is in the air on a skateboard}}     &
  \multicolumn{1}{|p{0.19\textwidth}|}{\footnotesize {\fontfamily{lmss}\selectfont \textbf{AoA}: A baby holding a remote control in its hand}}     &
  \multicolumn{1}{|p{0.19\textwidth}|}{\footnotesize {\fontfamily{lmss}\selectfont \textbf{AoA}: A man standing under a beach umbrella on top of a beach}}  &
  \multicolumn{1}{|p{0.19\textwidth}|}{\footnotesize {\fontfamily{lmss}\selectfont \textbf{AoA}: A woman holding a cell phone in her hand}} \\
  
    \multicolumn{1}{|p{0.19\textwidth}|}{\footnotesize {\fontfamily{lmss}\selectfont \textbf{Ours (UD)}: A horse is sitting in the grass near a lake}} &
  \multicolumn{1}{|p{0.19\textwidth}|}{\footnotesize {\fontfamily{lmss}\selectfont \textbf{Ours (UD)}: A man is doing a trick on a traffic light}}     &
  \multicolumn{1}{|p{0.19\textwidth}|}{\footnotesize {\fontfamily{lmss}\selectfont \textbf{Ours (UD)}: A baby is holding a cell phone in its mouth}}     &
  \multicolumn{1}{|p{0.19\textwidth}|}{\footnotesize {\fontfamily{lmss}\selectfont \textbf{Ours (UD)}: A man is standing on the beach with a surfboard}}  &
  \multicolumn{1}{|p{0.19\textwidth}|}{\footnotesize {\fontfamily{lmss}\selectfont \textbf{Ours (UD)}: A person is holding a pair of scissors}} \\
  
    \multicolumn{1}{|p{0.19\textwidth}|}{\footnotesize {\fontfamily{lmss}\selectfont \textbf{Ours (AoA)}: A horse running in a field near a body of water}} &
  \multicolumn{1}{|p{0.19\textwidth}|}{\footnotesize {\fontfamily{lmss}\selectfont \textbf{Ours (AoA)}: A man is jumping the air on a traffic light}}     &
  \multicolumn{1}{|p{0.19\textwidth}|}{\footnotesize {\fontfamily{lmss}\selectfont \textbf{Ours (AoA)}: A little girl holding a cell phone in her hand}}     &
  \multicolumn{1}{|p{0.19\textwidth}|}{\footnotesize {\fontfamily{lmss}\selectfont \textbf{Ours (AoA)}: A group of people standing on top of a beach}}  &
  \multicolumn{1}{|p{0.19\textwidth}|}{\footnotesize {\fontfamily{lmss}\selectfont \textbf{Ours (AoA)}: A woman is a pair of scissors in a brown}} \\
  
    \end{tabular}
    \endgroup
    \caption{Some qualitative samples from our baselines and Co-Occurence Updating models, referred as ours. }
    \label{fig:qualitative}
\end{figure*}

\subsection{Captioning images with uncommon object pairs}
To further investigate our proposed formulation, we provide Figure~\ref{fig:low_freq} for all of our models in~\tableref{row:13}-~\tableref{row:22}. In Figure~\ref{fig:low_freq}, we calculated the CHAIRs (Figure~\ref{fig:chairs}) and CHAIRi (Figure~\ref{fig:chairi}) for pairs of objects with low co-occurrence. For this we filtered those images of the MSCOCO dataset with pairs of objects with a co-occurrence lesser than 200. This accounts for 23.6\% of the MSCOCO test set. 
It can be recognized that original models UD (\tableref{row:13}) and AoA (\tableref{row:18}) have a much higher object bias on low frequency pairs than the other ones, an increase around 2\% for both models on CHAIRs and 0.2\%, 0.3\% on CHAIRi for UD and AoA. In addition, we see much better numbers on UD-L and AoA-L, so simple concatenation of labels lowers the object bias. Additionally, with the utilization of perfect labels (orange bars in Figure~\ref{fig:low_freq}), we appreciate that we obtain even better numbers on low frequency object pairs than the overall numbers calculated in Table~\ref{table:gt}. This suggests that our proposed augmentation can handle well on low frequency object pairs whether trained with cross entropy or self-critical. 

As well, we notice that the gap between the original models and Co-Occurrence Updating is bigger on the low frequency pairs. Therefore, our hypothesis on making co-occurrence matrix as uniform as possible making object bias lower holds valid. 

\subsection{Ablation Study}
\begin{table}[htp]
\centering
\begin{tabular}{c|ccc}
\toprule
 & \textbf{SS} & \multicolumn{1}{c}{\textbf{CHAIRs}~\cite{rohrbach2018object}} & \multicolumn{1}{c}{\textbf{CHAIRi}~\cite{rohrbach2018object}} \\ \midrule
\textbf{UD-L + Uni} & \xmark & 6.3 & 4.1 \\
\textbf{UD-L + Inv} & \xmark & 6.3 & 4 \\
\textbf{UD-L + Occ} & \xmark & 6.5 & 4.2 \\
\hline
\textbf{UD-L + Uni} & \cmark & 6.3 & 4 \\
\textbf{UD-L + Inv} & \cmark & 6.2 & 4 \\
\textbf{UD-L + Occ} & \cmark & \textbf{5.9} & \textbf{3.8} \\ \bottomrule
\end{tabular}
\caption{Ablation results on sentence simplification.}
\label{table:ablation}
\end{table}
Our final experimentation is on the analysis of sentence simplification. To see if the suggested formulation for sentence simplification has any effect on object bias, we decided to run UpDown model with and without sentence simplification. Our results can be found in Table~\ref{table:ablation}. 

As can be appreciated from the Table~\ref{table:ablation}, sentence simplification does not seem to have a lot of effect on Uniform and Inverse Multinomial sampling. Even though we always get better results on using sentence simplification, we only obtain 0.1\% which can be accounted for randomness. 

However, sentence simplification has a significant effect on Co-Occurrence Updating. Our conjecture regarding this phenomena is that since Co-Occurrence Updating selects more numbers of various pairs than the other two samplings, the models find a correlation between the adjectives and the replaced objects. As an example, usage of little or cute is usually adopted for boys or girls. When we replace the phrase ``cute little boy'' first with ``cute little broccoli'' and later with ``cute little clock''. The model will learn to associate ``cute little'' phrase first with broccoli and then with clock. However, in Uniform Sampling the models will merely discard this association because of the uniformity in nature and in Inverse Multinomial, only a handful of pairs will be associated with the phrase. Which is why we don't see a lot of disruption in Uniform and Inverse Multinomial.


\subsection{Qualitative Results}

Last but not least, we present some interesting qualitative samples in Figure~\ref{fig:qualitative}. Our first remark is that our models outperform the baselines in two ways, one of which is the deletion of hallucinated objects. This behaviour can be observed in the third and forth column where the baseline models predicted a surfboard, frisbee, beach umbrella or kite. These examples show the strong language prior that our models exploit. 

On the other hand, our models also outperform the baselines in that they not only delete the incorrect objects but also replace it with the correct one. As an example, in the first (or second) column of Figure~\ref{fig:qualitative}, while baseline models predict a dog (skateboard), our models corrects it to a horse (traffic light). One important remark is that sentences' verb or action prediction stays the same, \eg sitting, running, jumping, in which it calls for an augmentation technique for actions as well. 

Finally, we see that even in the case of wrongly produced captions (see fifth column), our models can still identify the correct object however they are constrained by the language models.

\section{Conclusion}
Since describing an image with a failure to correctly identify objects is not desirable to humans, we focus at the object bias in image captioning models. To reduce object hallucination in image captioning, we propose 3 different sampling techniques to augment sentences to be treated as ground truth to train image captioning models. By extensive analysis, we show that the proposed methods can significantly diminish our models' object bias on hallucination metrics. Also, we demonstrate that our methods can achieve much higher scores with the advances on object detectors. Moreover, we identify that our suggested techniques makes the models depend less on the visual features and by making co-occurrence statistics of objects uniform, and resulting in models generalizing better. But more importantly, we show that it is possible to decrease the object bias without needing extra data/annotations or increase in the model size or the architecture. Our hope is that this study incites more research on simple but effective methods to train deep models while keeping the model complexity untouched.

{\small
\bibliographystyle{ieee_fullname}
\bibliography{egbib}
}

\end{document}